\documentclass[twocolumn,letterpaper]{IEEEAerospaceCLS}
\usepackage[utf8]{inputenc}
\usepackage{graphicx}
\usepackage{amsmath}
\usepackage{amssymb}
\usepackage[style=ieee]{biblatex}
\usepackage{subcaption}
\bibliography{bib.bib}
\title{ieee-aero-2018}
\author{steven_morad }
\date{September 2018}

\begin{document}
\title{A Spring Propelled Extreme Environment Robot for Off-World Cave Exploration}
\author{%
Steven D. Morad\\
Space and Terrestrial Robotic Exploration\\
(SpaceTREx) Laboratory\\
Dept. of Aerospace and Mechanical Eng.\\
University of Arizona\\
smorad@email.arizona.edu
\and
Thomas Dailey\\
Space and Terrestrial Robotic Exploration\\
(SpaceTREx) Laboratory\\
Dept. of Aerospace and Mechanical Eng.\\
University of Arizona\\
daileytommy@email.arizona.edu
\and
Leonard Dean Vance\\
Space and Terrestrial Robotic Exploration\\
(SpaceTREx) Laboratory\\
Dept. of Aerospace and Mechanical Eng.\\
University of Arizona\\
ldvance@email.arizona.edu
\and
Jekan Thangavelautham\\
Space and Terrestrial Robotic Exploration\\
(SpaceTREx) Laboratory\\
Dept. of Aerospace and Mechanical Eng.\\
University of Arizona\\
jekan@email.arizona.edu}

\maketitle

\thispagestyle{plain}
\pagestyle{plain}

\maketitle

\begin{abstract}
    Pits on the Moon and Mars are intriguing geological formations that have yet to be explored. These geological formations can provide protection from harsh diurnal temperature variations, ionizing radiation, and meteorite impacts. Some have proposed that these underground formations are well-suited as human outposts. Some theorize that the Martian pits may harbor remnants of past life. Unfortunately, these geological formations have been off-limits to conventional wheeled rovers and lander systems due to their collapsed ceiling or "skylight" entrances. In this paper, a new low-cost method to explore these pits is presented using the Spring Propelled Extreme Environment Robot (SPEER). The SPEER consists of a launch system that flings disposable spherical microbots through skylights into the pits. The microbots are low-cost and composed of aluminium Al-6061 disposable spheres with an array of adapted COTS sensors and a solid rocket motor for soft landing. By moving most control authority to the launcher, the microbots become very simple, lightweight, and low-cost. We present a preliminary design of the microbots that can be built today using commercial components for under 500 USD. The microbots have a total mass of 1 kg, with more than 750 g available for a science instrument. In this paper, we present the design, dynamics and control, and operation of these microbots. This is followed by initial feasibility studies of the SPEER system by simulating exploration of a known Lunar pit in Mare Tranquillitatis. 
\end{abstract}

\tableofcontents

\section{Introduction}

\begin{table}[hp!]
    \centering
    \renewcommand{\arraystretch}{1.5}
    \begin{tabular}{c|c}
        Symbol & Meaning  \\
        \hline 
        $\pmb{Bold}$ & Matrix \\
        $\pmb{bold}$ & Vector \\
        $g$ & Lunar acceleration (1.625 m/s$^2$) \\
        $m$ & Bot mass (1 kg)\\
        $F$ & Force \\
        $J$ & Impulse \\
        $J_{net}$ & Net impulse\\
        $x$ & Launch spring displacement\\
        $t$ & Time \\
        $t_0$ & Time at launch \\
        $t_d$ & Time to move $d \pmb e_x$\\
        $t_f$ & Time of terrain impact \\
        $v_0$ & Launch speed \\
        $v_f$ & Speed just before impact \\
        $v_i$ & Impact speed \\
        $\theta$ & Launch angle \\
        $E : \{\pmb e_x, \pmb e_y, \pmb e_z\}$ & Inertial frame \\
        $B : \{\pmb b_x, \pmb b_y, \pmb b_z\}$ & Body-fixed bot frame \\
        $d$ & Horizontal distance to pit opening\\
        $h$ & Pit depth\\
        $\pmb R_0$ & Rotation matrix from $\pmb e_o$ to $\pmb e_d$ \\
        $\pmb R_1$ & Rotation matrix from $B$ to $E$ \\
        $\pmb e_o$ & Optimal impulse vector \\
        $\pmb e_t$ & Tangent vector to $\pmb e_d$ \\
        $\pmb e_d$ & Desired impulse vector \\
        $\phi$ & Angle between $\pmb e_o$ and $\pmb e_d$ \\
        $\phi_e$ & Error in $\phi$ \\
        $w_0$ & Angular speed of wheel 0\\
        $w_1$ & Angular speed of wheel 1\\
        $w_s$ & Angular speed for spin stabilization\\
        $w_i$ & Angular speed for impulse modulation\\
        $w_f$ & Net angular speed\\
        $w_e$ & Error angular speed\\
        $w_a$ & Actual angular speed\\
        $\pmb e_f$ & Unit vector for $w_f$\\
        $\pmb e_a$ & Unit vector for $w_a$\\
        $\pmb e_e$ & Unit vector for $w_e$\\
        $\Delta t_J$ & Impulse duration of thruster

    \end{tabular}
    \caption{Notation used in this paper}
    \label{tab:notation}
\end{table}





\begin{figure}[h]
    \centering
    \includegraphics[width=\linewidth]{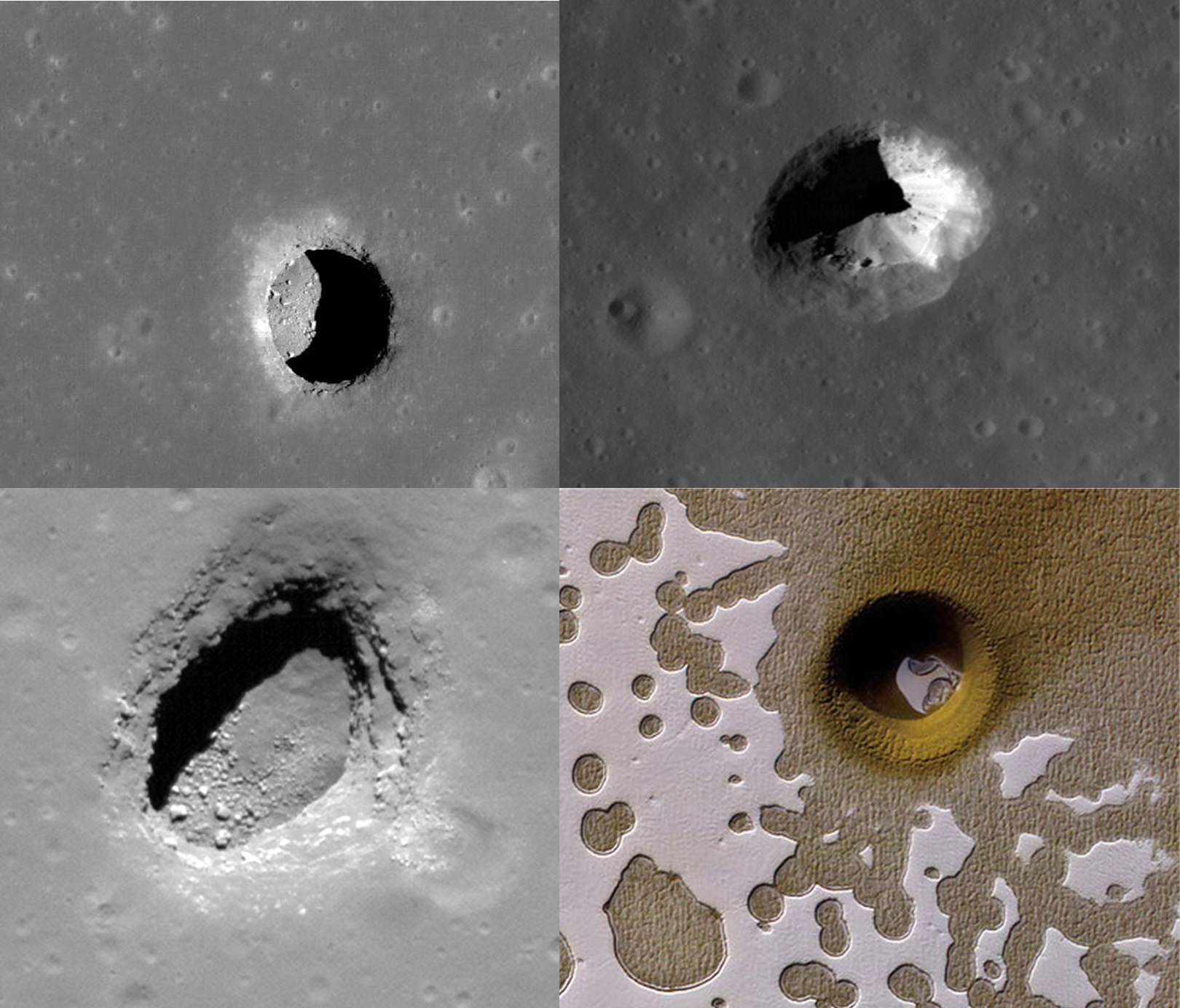}
    \caption{Lunar Reconnaissance Orbiter imagery of pits, (Top Left)  Mare Tranquillitatis, (Top Right) Mare Fecunditatis (Bottom Left) Mare Ingenii.  (Bottom Right) Example of a Martian pit taken by Mars Reconnaissance Orbiter \cite{nasa_photojournal_2010}. }
    \label{fig:pit}
\end{figure}

Pits on the Moon and Mars are geological mysteries that could provide valuable insight into past geo-history and shelter future humans (Fig. \ref{fig:pit}). Daga et al. discusses the reasoning for the exploration of these pits in a planetary science decadal survey report \cite{daga2009lunar}. These pits are one of the most promising locations for future research outputs, because they shield inhabitants from solar radiation, micrometeorites, and  temperature variations of hundreds of degrees \cite{WILLIAMS2017300}, \cite{coombs1992search}. On the Moon and Mars, some of these pits are believed to remnants of lava tubes. On the Moon, some of these pits are in polar permanently shadowed regions and may contain water-ice. On Mars, they offer protection from UV light, provide nearly constant temperatures, and potentially nutrient-rich volcanic regolith. This makes lava tubes one of the likely candidates for past life on Mars \cite{leveille2010lava}. These pits are relatively untouched by surface processes and are time capsules that can tell us about the early formation of the solar system.

Robots have not been sent inside an off-world pit, due to the difficulty and risk of vertically descending into a dark, unknown environment. To better understand if these pits harbor past-life or water, satellite observation is not sufficient \cite{daga2009lunar}. Missions are required that provide in-situ measurements.

Entering these pits consists of surviving a vertical drop on the order of one hundred meters \cite{cushing2007themis}. The floor near the opening is covered with rubble, causing mobility challenges. The ceiling entrances may not be structurally sound, and may collapse as a heavy rover drives up to the edge. Most pit exploration platforms fall into one of two categories: a tethered rappelling robot such as AXEL \cite{kerber2018moon}, or microbots \cite{doi:10.1063/1.1867276}, including the SphereX platform \cite{thanga2014spherex} \cite{raura2017spherical} \cite{kalita2017spherex} \cite{morad2018planning}.

Whittaker discusses both microbot and tethered robot approaches in his Nasa Innovative Advanced Concepts (NIAC) proposal on the robotic exploration of these pits \cite{whittaker2012technologies}. Tethered robots allow for a power and communications relay to be situated at the edge of the skylight, while a robot descends into the pit. JPL is currently pursuing this avenue with the AXEL and MoonDiver projects \cite{kerber2018moon}. There are concerns that the tether may cause issues. Upon ascent a tethered robot could get lodged under a rock, causing a mission failure. That being said, tethers are a versatile approach that allow for extended time in the pits because of their ability to transfer solar energy from the surface. Many recent robotic platforms for Lunar pit exploration have started to converge towards the tethered design.


Microbots offset the risks of one expensive rappelling robot by allowing multiple microbots to fail, without sacrificing the mission \cite{doi:10.1063/1.1867276}. Microbots would make their way to the opening of the pit, jump into the pit, and softly land using thrusters. Whittaker's concern with microbots is that they require extremely small components, that are decades away and may never actually exist. Designs like the SphereX microbot \cite{thanga2014spherex} \cite{raura2017spherical} \cite{kalita2017spherex} \cite{morad2018planning} require fast reaction wheels and multiple microthrusters to land inside the pit. While these components now exists thanks to wide availability of CubeSat components, they are still more expensive than initially envisioned. For these reasons, we set out to design a simple and low-cost microbot for Lunar and Martian pit exploration. Our intention is to have platforms of various costs to be tailored to specific missions.

\section{Design}
We propose the Spring Propelled Extreme Environment Robot (SPEER) system to develop a truly low-cost and disposable microbot. The SPEER system is a package consisting of a spring powered launcher and multiple microbot projectiles. The launcher uses a jack-in-the-box spring system to deploy one microbot at a time. Control happens before the bot leaves the launcher. A ballistic trajectory is computed and the bot is spin-stabilized by two wheels in the launcher before launch. 

\begin{figure}[t]
    \centering
    \includegraphics[width=\linewidth]{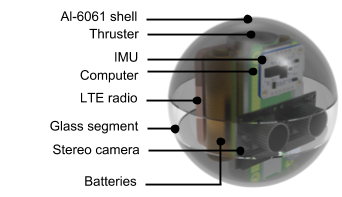}
    \caption{Rendering of the SPEER bot}
    \label{fig:cad0}
\end{figure}

The microbot projectile is a AL-6061 (aluminium) sphere, 4 cm in radius with a solid-rocket motor, IMU, batteries, camera, and LTE radio (Fig. \ref{fig:cad0}, Table \ref{tab:mass}). Other options were considered for propulsion including use of water-electrolysis propulsion \cite{pothamsetti2016electrolysis} and water steam propulsion \cite{rabade2017steam}.  Both provide additional advantages including increased control authority, however the water waste product makes it inappropriate for science mission where the focus is to find water ice in the pits.

The electronics are rated to survive the estimated -20 $^o$C environment of a Lunar pit \cite{colter}. Note that there is no reaction control system, which drastically cuts down on the mass of each microbot. The bot has room for science instruments, but other than that, it is a very simple system. By shifting all control to the launcher, large mass margins are afforded for science instruments and secondary payloads inside the bot.

\begin{table}[t]
    \centering
    \renewcommand{\arraystretch}{1.5}
    \begin{tabular}{c|c}
         \textbf{Component} & \textbf{Mass} (g) \\
         \hline
         2mm thick Al-6061 shell & 102 \\
         2mm thick borosilicate glass & 19\\
         Estes D12-3 20N-s solid rocket engine & 42 \\
         2$\times$Energizer L91VP 4.5 Wh lithium battery & 29 \\
         Raspberry-pi zero computer & 9 \\
         2$\times$Pi camera module & 6 \\
         Adafruit 6-axis IMU & 3 \\
         Sixfab 4G LTE radio & 18 \\
         Wiring & 3 \\
         \hline
         \textbf{Systems subtotal} & \textbf{231} \\
         Science payload & 769 \\
         \hline
         \textbf{Total} & \textbf{1000}
    \end{tabular}
    \caption{Mass breakdown for a SPEER bot}
    \label{tab:mass}
\end{table}

\begin{figure}[t]
    \centering
    \includegraphics[width=\linewidth]{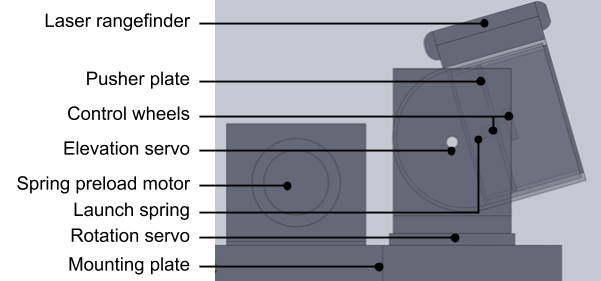}
    \caption{SPEER launcher diagram}
    \label{fig:launcher}
\end{figure}

\begin{table}[t]
    \centering
    \renewcommand{\arraystretch}{1.5}
    \begin{tabular}{c|c}
        \textbf{Component} & \textbf{Mass} (g) \\
        \hline
        $8 \times 8 \times 0.1$ cm aluminum pusher plate & 17\\
        Steel frame for pusher plate & 20\\
        PC105 4012 N/m spring & 5\\
        STP-MTRH stepper motor & 3800\\
        $2 \times$Pololu 1501MG high torque servo & 120\\
        $2 \times$ Parallax 120RPM continuous servo & 84\\
        1m of 0.25 inch steel cable & 160\\
        $2 \times$JSumo 13.25mm radius silicone wheel & 26\\
        Margin for mounting plates and brackets & 300\\
        Raspberry-pi zero computer & 9\\
        Lightware SF11 laser rangefinder & 35\\
        Pi camera module & 6\\
        \hline
        \textbf{Total} & \textbf{4592}
    \end{tabular}
    \caption{Mass breakdown for a SPEER launcher}
    \label{tab:my_label}
\end{table}

\begin{table}[t]
    \centering
    \renewcommand{\arraystretch}{1.5}
    \begin{tabular}{c|c}
        \textbf{Component} & \textbf{Energy} (mWh)  \\
        \hline
        2$\times$LV91VP Lithium battery & 9000\\
        \hline
        Raspberry-pi zero computer & 120\\
        2$\times$Pi camera module & 200\\
        Sixfab 4G LTE Radio & 1100\\
        Science Payload & variable\\
    \end{tabular}
    \caption{Power budget for a SPEER bot. Only components using a significant amount of energy are shown.}
    \label{tab:power_budget}
\end{table}

\begin{figure}[t]
    \centering
    \includegraphics[width=\linewidth]{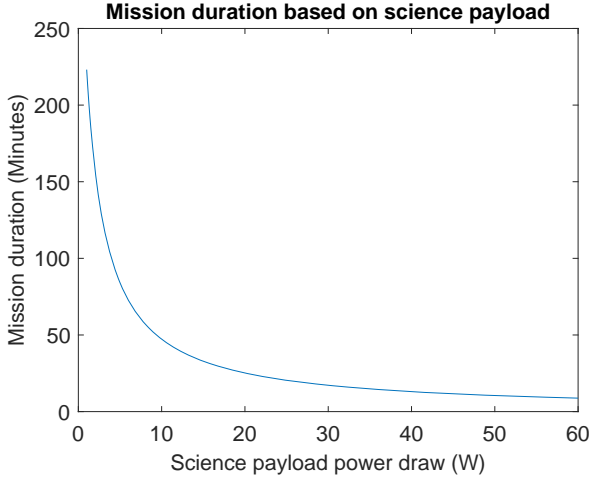}
    \caption{Mission duration based on the power draw of the science payload before losses.}
    \label{fig:power}
\end{figure}

\begin{figure}
    \centering
    \includegraphics[width=\linewidth]{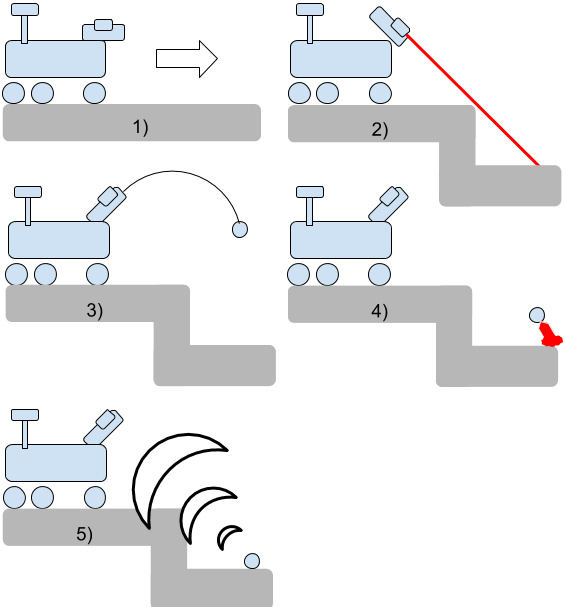}
    \caption{SPEER concept of Operations: 1) Drive to target. 2) Select landing site and find elevation using laser rangefinder, then compute trajectory. 3) Preload spring, apply spin, and launch. 4) Engine ignition and soft touchdown. 5) Utilize science instrument and relay data back to rover using radio. }
    \label{fig:conops}
\end{figure}

A rover would drive a safe distance from the edge of a pit and launch a SPEER bot. As the bot flies into the cave, it collects valuable terrain data by fusing its stereo camera and IMU measurements. The bot then uses its solid-thruster to safely land. Stereo imagery and video is streamed in real-time to the rover, in case landing fails. This ensures even if the bot is destroyed on impact, valuable imagery of the pit is broadcasted on time. After landing, the microbot can unpack and utilize the onboard science instrument and transmit findings back to Earth via the rover (Fig. \ref{fig:conops}).

\subsection{Spin Stabilization}
The Lunar pits are on the order of one hundred meters deep, so a free-falling bot will impact at around 18 m/s. This is too fast to survive, so a powered descent is required. The bot will need to orient its thruster to soft-land. In spacecraft orientation this is usually done using a control moment gyro or with three separate reaction wheels. CubeSat-grade gyros and reaction wheels are heavy, slow and expensive. 

Reaction wheels and gyros were not always available. Explorer-1 was the first satellite launched by the United States back in the 1950's but reaction wheels were not in use until the 1960's \cite{roberson1979two}. Explorer-1 utilized a method known as spin stabilization to orient itself. Spin stabilization is where a satellite spins along an axis to keep it pointed in a certain direction. The gyroscopic effect keeps the craft pointed toward its target, even with disturbances. We use spin stabilization to keep the SPEER oriented correctly.

\begin{figure}
    \centering
    \includegraphics[width=\linewidth]{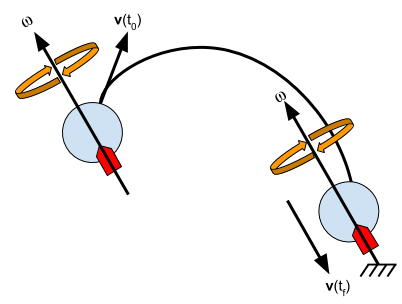}
    \caption{Spin stabilization of the SPEER. Angular velocity $\omega_f \pmb e_o$ is imparted at $t_0$ (launch) to ensure the thruster stays pointed towards $\pmb v(t_f)$. }
    \label{fig:rot}
\end{figure}

The SPEER launcher has a camera and laser rangefinder to find a landing site for the bot. Given the range and elevation of the landing site, we can compute the bot trajectory before launch (Eq. \ref{eqn:r}). We find $\pmb v(t_f)$, the velocity vector at the moment before impact (Eq. \ref{eqn:tf}). A spin imparted along $\pmb v(t_f)$ right before launch ensures the thruster stays pointed in the correct direction for a soft landing (Fig. \ref{fig:rot}, Eq. \ref{eqn:ws}). This removes the need for any reaction control system in the microbot itself.

\subsection{Soft Landing with a Solid-Fuel Engine}
Efficient soft landing is difficult. Electric CubeSat thrusters do not provide enough force to counteract gravity on the Moon or Mars. Liquid-fuelled CubeSat engines have a mass of over one kilogram, partially due to the heavy pressurized fuel container and flow control system. Some liquid-fuel engines also require additional power to heat or cool the propellant before use. Solid-fuel engines used in model rockets do not have these requirements and are therefore very simple and light. They consist of a propellant and oxidizer packaged in a tube that is ignited by running current through a resistor. 

\begin{figure}
    \centering
    \includegraphics[width=0.85\linewidth]{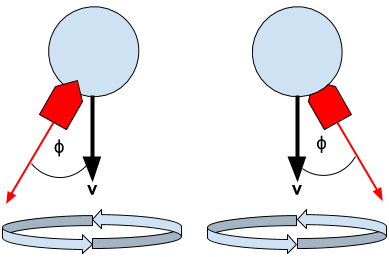}
    \caption{Adding precession allows modulating the net impulse of a solid-fuel rocket engine. Assume the bot is falling with velocity $\pmb v$ along $-\pmb e_z$. The thruster fires to soft-land. The precession causes the thruster to move in a circle about $v$. In the first instant, the thruster points at $\pi$. In the second instant, the thruster points at $2\pi$. The thrust components in the  plane orthogonal to $v$ cancel, reducing the net impulse. By varying angle $\phi$ the net impulse $J_{net}$ can be reduced to the desired impulse.}
    \label{fig:precession}
\end{figure}

The issue with using a solid-fuel engine is that they have a set impulse that cannot be changed. The engine cannot be easily shutdown and we do not know exactly how much $\Delta$v the SPEER bot will need before inspecting the cave. This would normally make solid-fuel engines unsuitable for soft-landing application, but in our case the thruster can be modulated by adding a torque-free precession to the bot using the launcher. The launcher uses powered wheels to create a precession by imparting a secondary angular velocity $w_0$, orthogonal to $\pmb v(t_f)$. This allows us to reduce the net impulse by pointing the thruster away from $\pmb v(t_f)$ (Fig. \ref{fig:precession}, Eq. \ref{eqn:w1}).

\section{Ballistics Analysis}
\begin{figure}
    \centering
    \includegraphics[width=\linewidth]{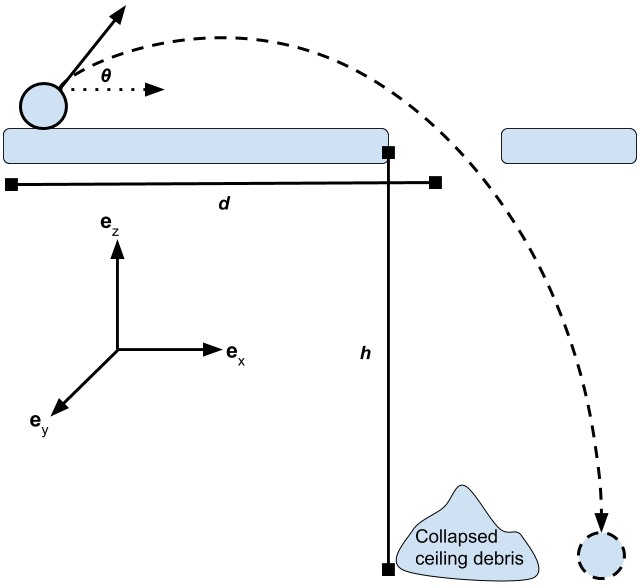}
    \caption{Visualization of SPEER ballistics}
    \label{fig:ic}
\end{figure}

We derive the equations of motion for the bot. Since the majority of the bot's flight is unpowered, it's motion is mostly ballistic with a soft-landing impulse $J_{net}$. The trajectory of the bot is illustrated in Fig. \ref{fig:ic}.

The bot leaves the launcher at time $t_0$ with launch velocity $v_0$ at angle $\theta$. The compression of the launch spring $x$ is computed as a function of $v_0$ using the following well established equation:
\begin{equation} \label{eqn:spring} x = v_0\sqrt{\frac{m}{k}} \end{equation}

The entire trajectory is in one plane so we can use planar dynamics. The acting forces are just gravity and an impulsive thrust.
\begin{equation} \pmb F = -m g \pmb e_z + \pmb f_{T} \delta(t_f - t) \end{equation}
\begin{equation} \pmb a(t) = -g \pmb e_z + \frac{\pmb f_{T} \delta(t_f - t) }{m} \end{equation}
Integrating we find
\begin{equation} \label{eqn:v} \pmb v(t) = \left(-gt + v_0 \sin\theta \right) \pmb e_z + v_0 \cos \theta \pmb e_x + \frac{\pmb J_{net}(t)}{m} \end{equation}

\begin{equation} \label{eqn:r} \pmb r(t) = \left( \frac{-gt^2}{2} + v_0 t \sin\theta \right) \pmb e_z + v_0 t \cos \theta \pmb e_x + \int \frac{\pmb J_{net}(t)}{m} dt \end{equation}

The impulse is instantaneous at $t_f$, so it can be ignored for most trajectory calculations. We compute the required launch velocity $v_0$ from the time to move $d \pmb e_x$, $t_d$
\begin{equation} \pmb r(t_d) \cdot \pmb e_z = 0 = t_d v_0 \sin{\theta} - g\frac{t^2}{2} \end{equation}
\begin{equation} t_d = \frac{2v_0 \sin{\theta}}{g} \end{equation}

\begin{equation} \pmb r(t_d) \cdot \pmb e_x = d = v_0 \cos{\theta} t_d \implies d = v_0 \cos{\theta} \frac{2v_0 \sin{\theta}}{g} \end{equation}
\begin{equation} v_0 = \sqrt{\frac{dg}{2 \sin{\theta} \cos{\theta}}} \end{equation}

We can compute the velocity right before the instantaneous impulse, $v_f = || \pmb v(t_f) ||$

\begin{equation} \pmb r(t_f) \cdot \pmb e_z = -h = v_0 \sin{\theta} t_f - g\frac{t_f^2}{2} \end{equation}
\begin{equation} \label{eqn:tf} t_f = \frac{1}{g} \left( \sqrt{2gh + v_0^2 \sin^2{\theta}} + v_0 \sin{\theta} \right) \end{equation}

Plugging $t_f$ into Eq. \ref{eqn:v}
\begin{equation} \pmb v(t_f) = v_0 \cos{\theta} \pmb e_x - \sqrt{2gh + v_0^2 \sin^2{\theta}} \pmb e_z \end{equation}

\begin{equation} v_f = || \pmb v(t_f) ||= \sqrt{v_0^2 + 2gh} \end{equation}

\section{Rotational Analysis}
Now that the ballistics equations are derived, we analyze the rotation required for a soft landing. We must ensure the thruster is pointed correctly, resulting in the impulse $J_{net}$ cancelling out the velocity such that $v(t_f) = 0$. The mass of the bot is distributed in such a way that the center of mass is in the center of the sphere. The thrust vector will always point through the center of mass, which means the thrust will not apply a moment or affect the rotation.

We define the negative of our optimal impulse vector in the inertial frame as 
\begin{equation} \pmb e_o = \frac{\pmb v(t_{f})}{|| \pmb v(t_{f}) ||} \end{equation} and attach a body-fixed frame $B : {\pmb b_x, \pmb b_y, \pmb b_z}$ to the bot such that the thruster points along the $-\pmb b_z$ axis. Two wheels, $\{0,1\}$ are used to orient the bot and apply angular velocities to the bot.

\subsection{Applying Spin Stabilization}
We want the bot to be robust against noise that would point the thruster ($- \pmb b_z$) away from $\pmb e_o$, so we spin stabilize about $\pmb e_o$. We use $\pmb R_0$ to align $- \pmb b_z$ with $\pmb e_o$

\begin{equation} \pmb R_0 = \begin{bmatrix} 
\cos \theta & 0 & -\sin \theta \\
0 & 1 & 0 \\
\sin \theta & 0 & \cos \theta
\end{bmatrix} \end{equation}

Now that $\pmb e_o$ is aligned with $- \pmb b_z$ We can spin wheel 0 at $w_0$ to impart an angular velocity of $-w_{s} \pmb b_z$ to the bot. Let $r_w$ be the radius of the wheel and $r_b$ be the radius of the bot. Assuming no-slip between the wheel and bot, we have

\begin{equation}
    \label{eqn:ws}
    \pmb w_{0} =  \frac{w_{s} \pmb b_z r_b}{r_w}
\end{equation}

Using Eq. \ref{eqn:ws}, we show how to apply a spin stabilization of $w_s \pmb b_z$ to keep the thruster pointing towards $\pmb e_o$ (Fig. \ref{fig:wheel}).

\subsection{Applying a Precession}

Because of the nature of solid-fuel thrusters, we want to be able to modulate our net impulse by creating a torque-free precession of angle $\phi$ about $\pmb e_o$ (Fig. \ref{fig:precession}). $J_{net}$ is our desired impulse while $J$ is the impulse rating of the thruster.

\begin{equation} J_{net} = J \cos{\phi} = f_{T} \delta(t_f - t) \cos \phi \end{equation}
\begin{equation}  \label{eqn:jnet} \phi = \arccos{\frac{J_{net}}{J}} \end{equation}

With $\phi$, we construct rotation matrix $\pmb R_1$. We can use this to compute our desired impulse magnitude $J_{net}$ and desired impulse vector $\pmb e_d$

\begin{equation} \pmb R_1 = \begin{bmatrix} 
\cos \phi & 0 & -\sin \phi\\
 0 & 1 & 0\\
\sin \phi & 0 & \cos \phi
\end{bmatrix} \end{equation}

\begin{equation} \pmb e_d = \pmb R_1 \pmb e_o \end{equation}

After wheel 0 has spin stabilized the bot, wheel 1 rotates at $w_1$ along $\pmb b_y$ to create a precession $w_i$ to modulate the thruster impulse. 

\begin{equation}
    \label{eqn:w1}
    \pmb w_{1} =  \frac{- w_{i} \pmb b_y r_b}{r_w}
\end{equation}

Eq. \ref{eqn:w1} tells us how fast to spin wheel 1 to apply a precession.



\subsection{Nulling the Horizontal Components}

To ensure that the horizontal components of the velocity are nulled, we need to make sure that the thruster completes full rotations over the impulse duration. One can imagine a case where the entire impulse happens over one quarter of a rotation, which would produce a nonzero impulse in the $\pmb b_x, \pmb b_y$ plane. We will find an expression for the magnitude $w_f$, such that the impulse happens over a multiple of $2\pi n$.

To complete $n$ full rotations over a given impulse duration $\Delta t_J$ we have
\begin{equation} \frac{w_f}{\Delta t_J} = 2\pi n \end{equation}
So the angular velocity magnitude $w_f$ is constrained by the impulse duration
\begin{equation} w_f = 2\pi n \Delta t_J \end{equation}
A thruster that burns over one second would require $w_f$ of $2\pi, 4\pi, 6\pi...n \text{ rad/s}$.

\subsection{The Effect of Noise on Impact Velocity}
We generally want to pick the largest $n$ we can. This reduces the divergence between the desired thruster direction $\pmb e_d$ and the noise present in the actual thruster direction $\pmb e_a$, ultimately reducing the impact velocity. We analyze how spin stabilization reduces the impact velocity $v_i$, to make for a softer landing. Let the actual spin $\pmb w_a$ be a sum of the desired spin $\pmb w_f$ and some error $\pmb w_e$.

\begin{equation} \pmb w_a = w_f \pmb e_d + w_e \pmb e_t \end{equation}

\begin{figure}
    \centering
    \includegraphics[width=\linewidth]{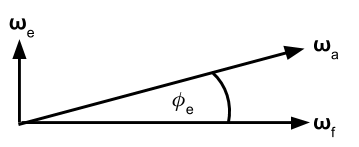}
    \caption{Geometric result of noise vector $\pmb w_e$ on the actual axis of rotation $\pmb w_a$}
    \label{fig:sum}
\end{figure}

We can express the $\pmb w$ vectors geometrically with the resulting rotation vector $w_a$ deviating from $\pmb e_d$ by angle $\phi_e$ (Fig. \ref{fig:sum}).

We can find $\phi_e$ using the definition of the dot product
\begin{equation} \phi_e = \arccos{\left(\frac{\pmb w_f \cdot \pmb w_a}{||\pmb w_f|| ||\pmb w_a||}\right)} \end{equation}

Thus, we can see how the optimal thruster direction $\pmb w_f$ changes to $\pmb w_a$ with noise $\pmb w_e$. 

We show exactly how this noise effects impact velocity $v_i$. Let $\pmb e_f$, $\pmb e_a$, and $\pmb e_e$ be the unit vectors for the respective $\pmb w$'s, with $J_{net}$ being the instantaneous thruster impulse. Then we have

\begin{equation} \pmb v_i = v_f \pmb e_f - J_{net}(\cos{\phi_e} \pmb e_f + \sin{\phi_e} \pmb e_e ) \end{equation}
We want zero landing velocity so $J_{net} = v_f$
\begin{equation} \pmb v_i = v_f (1 - \cos{\phi_e}) \pmb e_f + v_f \sin{\phi_e} \pmb e_e \end{equation}
We find the impact magnitude as
\begin{equation} || \pmb v_i || = v_i = v_f \sqrt{(1 - \cos{\phi_e})^2 + \sin^2{\phi_e}} \end{equation}
\begin{equation} v_i = v_f \sqrt{1 - 2\cos{\phi_e} + \cos^2{\phi_e} + \sin^2{\phi_e}} \end{equation}
\begin{equation} v_i = v_f \sqrt{2 - 2\cos{\phi_e}} \end{equation}

\subsection{Putting it all Together}
\begin{figure}
    \centering
    \includegraphics[width=\linewidth]{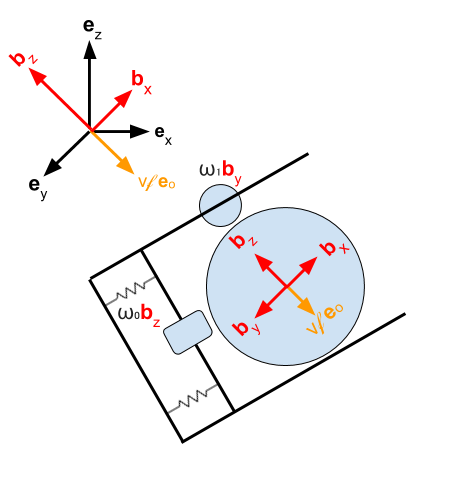}
    \caption{The wheels applying angular velocity to the bot. The body-fixed frame is in red, with the inertial frame in black. $\pmb b_z$ is parallel to $\pmb e_o$.}
    \label{fig:wheel}
\end{figure}

With spin stabilization component $\pmb w_s$ and impulse modulation component $\pmb w_i$, we can find an equation for the total angular velocity of the bot $\pmb w_f$ (Fig. \ref{fig:wheel}).

\begin{equation}
    w_f \pmb e_d = w_s \pmb b_z + w_i \pmb b_y
\end{equation}

\section{Discussion}
One possible application of the SPEER robots is to probe lunar pits to determine if ice resides in them. As the temperature in the pits remains stable at around -20 $^o$C, the conditions could be ideal for ice accumulation. To determine if there is ice in these pits, each SPEER would need to carry a science payload such as a compact  neutron generator and detector.  The other possibility is to just a carry a neutron generator and have the neutron detector on the rover.  This would require line-of-sight between the rover and SPEER.

\subsection{Science Payload}
Large pieces of ice can be identified using the cameras, but that is not the case for smaller ice crystals dispersed in the regolith. NASA's curiosity rover carried an experiment called Dynamic Albedo of Neutrons (DAN). This experiment used a 14 MeV neutron generator to detect water and hydrated minerals up to 1m below the surface. Sandia National Labs recently designed a compact neutron generator, called the "neutristor". The neutristor was designed to be placed near tumors to allow cancer patients to continue their radiation therapy at home. As such, the neutristor was built to be powered by a single battery, with a compact 1.5 $\times$ 3cm form factor \cite{elizondo2012surface}. 

Like the DAN experiment, the neutristor also has an output energy of 14 MeV. Due to its low power usage and small size, it can easily fit inside a SPEER bot.  In addition, the SPEER bot would ideally carry the neutron detector. Like the bots, the neutristor is disposable, with an operational time of 1000 seconds. Multiple bots can be launched to different places in the pit to find the areas with the highest concentration of ice or hydrated minerals.

\subsection{Dynamics}
\begin{figure}[h]
    \begin{subfigure}[t]{0.45\textwidth}
        \includegraphics[width=\linewidth]{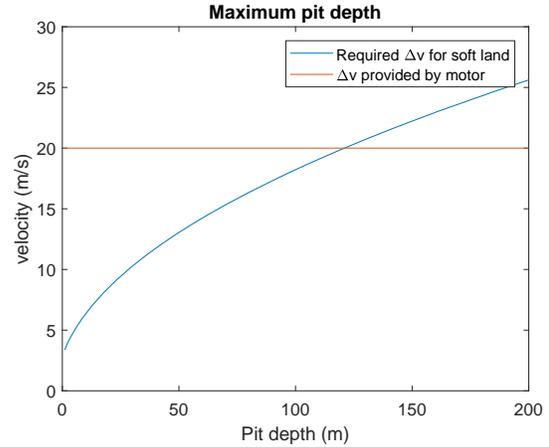}
        \subcaption{The maximum depth of a pit is limited by the total impulse. This plot uses parameters: $d=5$, $\theta=45 \deg$}
        \label{fig:depth}
    \end{subfigure}
    \begin{subfigure}[t]{0.45\textwidth}
        \centering
        \includegraphics[width=\linewidth]{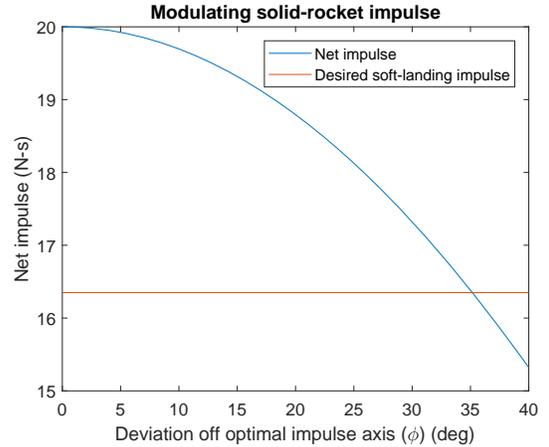}
        \subcaption{Precession allows the throttling of a solid-fuel engine. For an 80m depth pit, 35 degrees of precession results in a soft-landing. }
        \label{fig:mod}
    \end{subfigure}
    \caption{Soft landing requirements}
\end{figure}

Using the equations presented in the analysis, we can compute flight parameters for the pit discovered by Haruyama et al. Haruyama estimates the depth of the pit in the Mare Tranquillitatis as 80m \cite{haruyama2009possible}. We want to throw the bot in from distance $d \pmb e_x$ so that the rover doesn't drive up to the unstable opening. We assume the bot is launched $d=5$m from the cave opening, and that the depth of the cave is $h=80$m (Fig. \ref{fig:ic}). Optimizing for maximum depth, the bot can soft-land after drops as large as 123.5m. For deeper pits, 14g of the science payload mass can be allocated for the larger 30 N-s thruster variant (Fig. \ref{fig:depth}).

\begin{figure}
\centering
    \begin{subfigure}[t]{0.45\textwidth}
        \includegraphics[width=\linewidth]{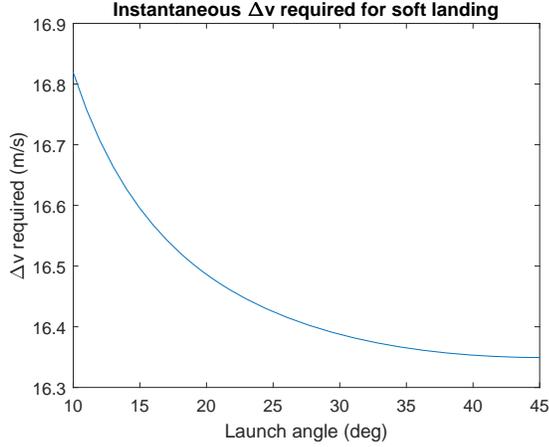}
        \subcaption{Soft landing $\Delta$v requirements for various launch angles $\theta$ where $d=5$m and $h=80$m.}
        \label{fig:inst_dv}
    \end{subfigure}
    \begin{subfigure}[t]{0.45\textwidth}
        \includegraphics[width=\linewidth]{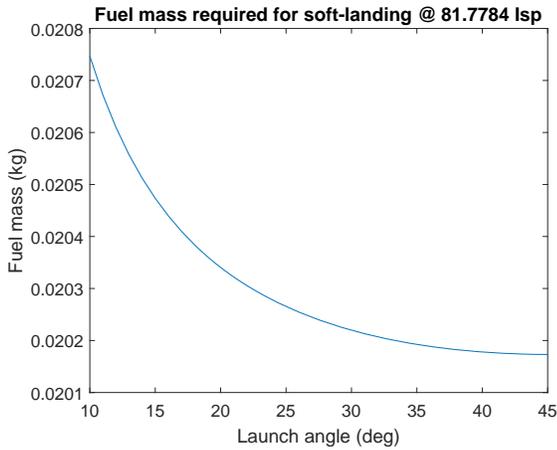}
        \subcaption{Soft landing fuel requirements for various launch angles $\theta$. The Estes D12-5 solid-fuel rocket engine contains 24.93g of fuel at 81.7784 $I_{sp}$.}
        \label{fig:fuel}
    \end{subfigure}
    \caption{Criteria used to select a value of $\theta$}
    \label{fig:bork}
\end{figure}

We look at various launch angles $\theta$. Since $d$ is small, the difference in final velocities $v_f$ is also small (Fig. \ref{fig:inst_dv}). This translates to a small change in the fuel required (Fig. \ref{fig:fuel}).

\begin{figure}
        \includegraphics[width=\linewidth]{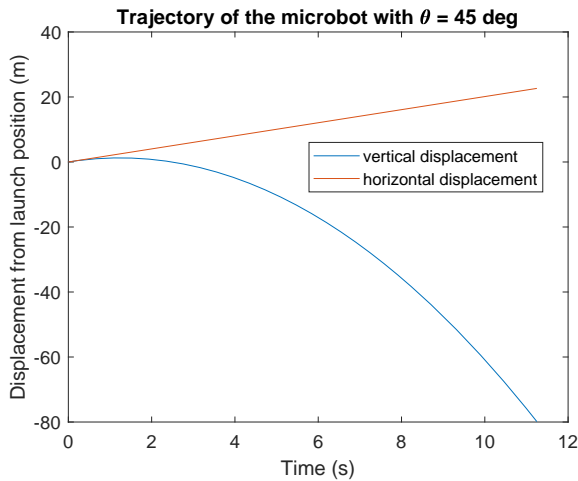}
    \caption{Ideal trajectory with $v_i=0$ and how $v_i$ varies with rotational noise}
 \label{fig:traj}
\end{figure}

In a typical scenario, we select a launch angle of $45 \deg$ and compute the trajectory of the bot using Eq. \ref{eqn:r}, plotted in Fig. \ref{fig:traj}. This results in a desired launch velocity $v_0=2.8460$ m/s. Using spring stiffness $k=4012$ N/m, we use Eq. \ref{eqn:spring} to compress the spring $x=4.49$cm to obtain our launch velocity.

\section{Conclusion}

We have proposed the SPEER system, which is made entirely of commercial components. The SPEER microbot costs under 500 USD each, with a total mass of 1 kg including up to 770 g of science payload.  The components selected are capable of surviving the modest temperatures and low-radiation conditions of a Lunar or Martian pit. We have simulated a deployment of the SPEER architecture for exploration of a Lunar pit in the Marius hills and have provided numerical solutions for a soft touchdown. We have made some simplifying assumptions, but the values show that this is a feasible approach to off-world pit exploration.

SPEER better addresses the problem of microbot designs relying on small components that do not yet exist or are expensive. Certain components, like reaction wheels and liquid-fuel thrusters face reliability challenges due to miniaturization. We remove the need for the reaction wheels by externally spin stabilizing the bots. We add a second spin for impulse control, removing the need for a liquid-fuel thruster. This transfer of control authority from the microbot to the launcher allows for significant mass savings. 

In-situ measurements provide a better picture of the inside of these pits than is possible with state-of-the-art recon satellites such as the Lunar Reconnaissance Orbiter (LRO). Lunar and Martian pits provide protection from radiation and diurnal temperature variations. On the Moon, they may contain water ice. On Mars, they may hold remnants of past life. Until we send robots inside them, we will not know. The SPEER system is a cost-effective, low-risk pathway for us to get a first look inside these off-world pits.




\section{Future Work}
The details of how the SPEER bots are packaged on the rover is not discussed. Some ideas are to place three of them to be packaged in an internally modified 3U PPOD, which would act like a magazine on a firearm. 


We are working on building a SPEER prototype. We plan to experimentally test the performance of SPEER in the rugged environments of Arizona and New Mexico. If testing goes well, we plan to look into operating a SPEER system in a practical capacity on Earth. Mapping abandoned mine shafts too small for quadcopters or squeezing through tight crevasses to find trapped mountain climbers are just two Earth-based applications of the SPEER.

\printbibliography

\thebiography
\begin{biographywithpic}
{Steven D. Morad}{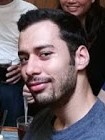}
is an aerospace engineering PhD student at the University of Arizona. He is currently a member of the Space and Terrestrial Robotic Exploration (SpaceTREx) Laboratory at the University of Arizona. Steven received his bachelor's in computer science at the University of California, Santa Cruz. He previously worked on operating systems and automation at Facebook. Steven is currently interested in autonomy for extreme off-world environments, such as Lunar pits and the surface of icy moons.
\end{biographywithpic} 
\begin{biographywithpic}
{Thomas Dailey}{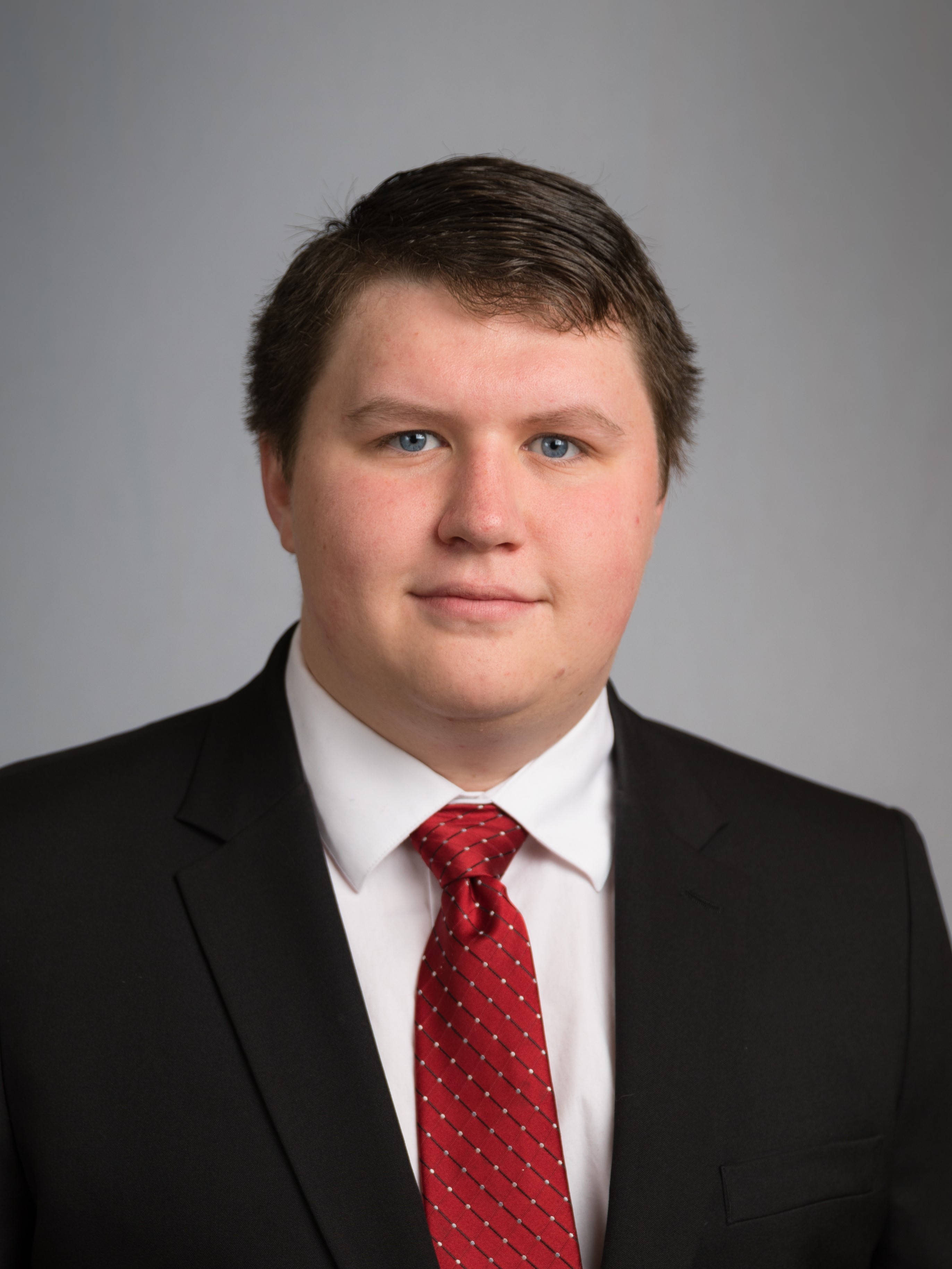}
is an undergraduate student in systems and industrial engineering at the University of Arizona. He is an undergraduate researcher for the Space and Terrestrial Robotic Exploration (SpaceTREx) Lab. Thomas is interested in spacecraft design and shape memory alloys.
\end{biographywithpic} 
\begin{biographywithpic}
{Leonard Vance}{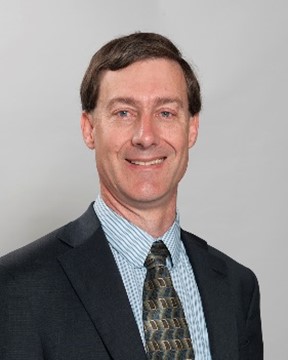}
is a PhD Student in Aerospace and Mechanical Engineering at the University of Arizona, and a Retired Senior Engineering Fellow from Raytheon Missile Systems.  His experience over the course of 33 years includes concept definition and development for a variety of leading edge space systems for both defense and exploratory purposes.
\end{biographywithpic}
\begin{biographywithpic}
{Jekan Thangavelautham}{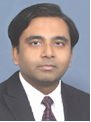}
has a background in aerospace engineering from the University of Toronto. He worked on Canadarm, Canadarm 2 and the DARPA Orbital Express missions at MDA Space Missions.  Jekan obtained his Ph.D. in space robotics at the University of Toronto Institute for Aerospace Studies (UTIAS) and did his postdoctoral training at MIT's Field and Space Robotics Laboratory (FSRL).  Jekan is an assistant professor and heads the Space and Terrestrial Robotic Exploration (SpaceTREx) Laboratory at the University of Arizona.  He is the Engineering Principal Investigator on the AOSAT I CubeSat Centrifuge mission and is a Co-Investigator on SWIMSat, an Airforce CubeSat mission concept to monitor space threats.
\end{biographywithpic} 

\end{document}